\documentclass[letterpaper, 10 pt, conference]{ieeeconf}
\usepackage[utf8]{inputenc}
\IEEEoverridecommandlockouts
\DeclareUnicodeCharacter{2212}{-}
\pdfoutput=1 

\usepackage[british]{babel}

\usepackage[T1]{fontenc}
\usepackage{newtxtext}

\usepackage{graphicx}
\graphicspath{{figures/}}

\usepackage{amsmath,amssymb,amsfonts}

\usepackage[hyphens]{url}
\usepackage[colorlinks=true, allcolors=blue]{hyperref}
\usepackage{balance}

\title{\LARGE \bf
LoKiTO: Local Kinodynamic Trajectory Optimization for UAVs based on Memory-Efficient Neural Distance Fields}
\author{
Guillermo Gil$^{1}$, Jose Antonio Cobano$^{1}$, Luis Merino, \textit{Member, IEEE}$^{1}$ and Fernando Caballero$^{1}$%
\thanks{$^{1}$Service Robotics Laboratory -- Universidad Pablo de Olavide (Seville), Spain
{\tt\small \{ggilgar, jacobano, lmercab, fcaballero\}@upo.es}}
}

\begin{document}

\maketitle
\thispagestyle{empty}
\pagestyle{empty}

\begin{abstract}

This paper introduces LoKiTO, a 3D local kinodynamic trajectory optimizer that combines a Chebyshev polynomial parameterization with an online neural Euclidean Signed Distance Field (ESDF) representation for aerial robotics. While current gradient-based planners often struggle with the high memory requirements of explicit volumetric maps or fail to strictly enforce dynamic limits, LoKiTO embeds kinodynamic constraints directly into a modular cost function and exploits the numerical conditioning properties of the Chebyshev basis. Collision awareness is provided by a continuous, memory-efficient neural ESDF updated online from LiDAR measurements, which drastically reduces map memory consumption to approximately 1.0 MB. Benchmarking in highly cluttered simulated environments against state-of-the-art baselines (Fast-Planner and EGO-Planner) demonstrates the superiority of our approach. Experimental results show that LoKiTO achieves over 99.4\% dynamic feasibility compliance, significantly outperforming the baselines, and maintaining higher median clearance from obstacles. By coupling a well-conditioned continuous parameterization with explicit kinodynamic reasoning, LoKiTO provides a practical and extensible basis for local replanning in aerial robotics. The source code is open source and can be found at: \url{https://anonymous.4open.science/r/LoKiTO-RAL} 

\end{abstract}

\section{Introduction} \label{sec:intro}

Aerial robotics, and autonomous drones in particular, is a rapidly growing field of robotics thanks to its wide range of applications. Drones can perform tasks efficiently while avoiding endangering human lives in the process \cite{Ullah2023}. However, scenarios of critical tasks such as disaster management \cite{MOHDDAUD2022} or urban firefighting \cite{fr21mbzirc} are dynamic and partially unknown. Because of that, trajectory replanning has become a critical part of any drone's computation pipeline. This is clearly shown by the increasing investigation effort in that direction in the last decade \cite{zhou2025}. The creation and management of an up-to-date 3D map of the surrounding environment also becomes necessary in order to avoid and maintain a safe distance from obstacles. % Our paper describes a framework for aerial drones that combines both a way to leverage the drone sensors to create a neurally-generated 3D continuous distance field of its surroundings, and a planner that outputs continuous smooth local polynomial trajectories over that map.

Space representation and mapping has seen significant change and evolution, from the earliest occupancy grids, through triangular meshes and point cloud representations, to modern methods such as Gaussian splatting \cite{fei2025}, Gaussian EDFs \cite{maese2026gedfloc} or bitmask-based TSDFs \cite{maese2026dbtsdf}. Each of these paradigms offers different trade-offs between memory consumption, query efficiency, and geometric fidelity of captured surfaces, without a clear winning option for general usage among them \cite{deng2026}.

Among these alternatives, Euclidean Signed Distance Fields (ESDFs) stand out. An ESDF encodes, at every point in space, the signed Euclidean distance to the closest obstacle. The resulting scalar field is Lipschitz-continuous and differentiable everywhere except at the cut-locus \cite{Jones:TVCG2006}. The gradient's norm is also, by definition, equal to one. These properties have made ESDFs particularly attractive for robot navigation \cite{Oleynikova2016}. 
In practice, ESDFs are most commonly built on discrete voxel grids that are updated incrementally as new sensor measurements arrive, as exemplified by Voxblox \cite{voxblox_2017}, FIESTA \cite{FIESTA}, or Voxfield \cite{pan2022}. Neural representations eliminate some well-known limitations of discrete counterparts: the trade-off between resolution and memory, and the need to interpolate between neighboring cells whenever a distance or gradient is required \cite{Ortiz:etal:iSDF2022}.
% Recent works have explored the use of neural networks as implicit, continuous parameterizations of a distance field. iSDF \cite{Ortiz:etal:iSDF2022} learns a neural SDF online from posed depth images, providing smooth and continuously differentiable queries at arbitrary locations, while HIO-SDF \cite{vasilopoulos_hiosdf_2024} combines a coarse explicit grid with a neural component to scale incremental mapping to larger workspaces. Further recent contributions have refined this line of work, including super-lightweight neural mapping for large-scale scenes \cite{shi2026} and loss functions specifically tailored to mitigate spurious extrema in neural SDF reconstruction \cite{li2025}. A related family of methods replaces the spatial distance with the arrival time predicted by a physics-informed neural network: Neural Time Fields (NTFields) \cite{liu2025_ntf} provide differentiable and continuous fields that can be directly exploited for motion planning.

The appeal of neural ESDFs is best understood when looking at how they are exploited inside a trajectory planner. Sampling- and search-based methods produce geometrically optimal paths but disregard flight safety and dynamic feasibility, requiring post-processing before deployment. Meanwhile, distance fields provide the gradient information that turns planning into a smooth optimization problem \cite{Oleynikova2016}.  %While traditional approaches (sampling- or searching-based) generate optimal paths without taking flight safety into account, making them unsuitable for real-world applications without post-processing. A wide set of state-of-the-art works shows that ESDFs inherently solve this safety issue, resulting in very convenient tools for real-life applications that compute safe, feasible, and fast paths \cite{iros2022_edf, jacquet2025, Tordesillas_2019}. However, existing work focuses mainly on discretized mapping.

% This paper focuses on performing 3D local continuous trajectory planning on an on-line neurally-generated ESDF. The system gathers information through an onboard 3D LiDAR to train a neural network using an efficiency- and speed-oriented method. Then, the most up-to-date version of the network is integrated into an iterative non-linear trajectory optimizer that accounts for a wide range of kinodynamic constraints.

We present a local kinodynamic trajectory optimizer that exploits the properties of both a continuously updated neural ESDF and a polynomial basis to achieve replanning rates suitable for online and local operation. We improve safety and flight time of the resulting trajectories by integrating drone dynamics in the optimization. The proposed parameterization also makes the integration of desired constraints straightforward and the modular structure of the cost function allows the planner to be easily extended with additional task-specific requirements.

This paper is organized into six sections. Section \ref{sec:relat} discusses the state of the art. Section \ref{sec:frame} introduces the overall framework structure, and Section \ref{sec:path} delves into the design of the polynomial trajectory planner. Section \ref{sec:exper} validates the approach experimentally, including a comparison with other state-of-the-art planners, and Section \ref{sec:conc} presents the conclusions and future development lines.

\section{Related Work} \label{sec:relat}

Recent approaches to 3D trajectory planning can be broadly grouped into two main families based on the chosen optimization philosophy. Hard-constrained methods formulate trajectory generation as a constrained optimization problem in which dynamic feasibility and obstacle avoidance are encoded as explicit constraints. The canonical example is the minimum-snap formulation, which generates smooth polynomial trajectories by minimizing high-order derivatives subject to waypoint and continuity constraints \cite{mellinger2011}. Building on this idea, \cite{Ding_ICRA_2018} combines a front-end kinodynamic search with a back-end elastic optimization, while \cite{Tordesillas_2019} casts the problem as a mixed-integer quadratic program whose binary variables select among convex safe regions, providing rigorous safety guarantees at a notably higher computational cost.

The second family, gradient-based optimization methods, relaxes hard constraints into soft penalty terms within a differentiable cost function minimized by local gradient descent. CHOMP \cite{chomp} popularized this paradigm by jointly optimizing smoothness and obstacle cost through functional gradients. Most quadrotor planners in this line rely on an ESDF to supply the obstacle gradient \cite{Zhou2019}, although maintaining the field is expensive relative to the fraction of it that the optimizer actually queries. EGO-Planner \cite{ego-planner} therefore avoids ESDFs, computing repulsive directions directly from colliding trajectory segments, although gradients remain valid only near the current solution.

% The other main current is gradient-based optimization methods, which relax hard constraints into soft penalty terms in a differentiable cost function, which is then minimized via local gradient descent. The CHOMP framework \cite{chomp} popularized this paradigm by jointly optimizing smoothness and obstacle cost using functional gradients. Many subsequent quadrotor planners rely on an ESDF to provide the obstacle gradient: in \cite{Zhou2019} a B-spline representation is used together with an ESDF to enable robust and efficient replanning at high speeds, and the MIQP-based planner of \cite{Tordesillas_2019} exploits distance information for safe corridor construction. There are other works that avoid the cost of maintaining a full ESDF. For example, EGO-Planner \cite{ego-planner} computes obstacle gradients directly from occupied trajectory segments. Recently, further research has been done in that direction: in \cite{Fu_2024} motion prediction is combined with gradient-based optimization for online replanning around moving obstacles, while the authors of \cite{akhihiero2025} integrate trajectory optimization with vector-field guidance, using the field as an initial reference and subsequently refining it for smoothness and obstacle avoidance.

A second axis concerns how the trajectory itself is parameterized, which determines the cost of enforcing constraints and the coupling between geometry and timing. For piecewise polynomials, kinodynamic limits and boundary conditions translate into algebraic expressions of the coefficients. On the other hand, B-splines enforce feasibility through the convex-hull property of their control points \cite{Ding_ICRA_2018, Zhou2019}, which is conservative and leaves time allocation to an outer heuristic, whereas recent formulations couple the spatial and temporal profiles within a single sparse representation at the cost of a specialized back-end \cite{wang2022minco}.

%In contrast, parameterizing the trajectory with Chebyshev polynomials offers a distinct advantage in the context of dynamic constraints. While the dynamic constraints become non-linear in the Chebyshev coefficients, the resulting optimization problem is substantially smaller than its B-spline counterpart, and the orthogonality of the basis promotes faster convergence. B-splines, on the other hand, provide formal guarantees of constraint satisfaction through the convex hull property, but this comes at the cost of a larger number of decision variables. More critically, even when B-splines are employed, jerk constraints remain fundamentally non-convex and computationally expensive to handle, often requiring specialized reformulation techniques to achieve convexity \cite{Ji2024Convex}. This additional complexity further motivates our choice of a compact polynomial representation with an orthogonal basis.

A third axis is how environmental information is encoded. Some early continuous-time formulations optimize a polynomial trajectory against an ESDF cost for online replanning \cite{Oleynikova2016}, while EWOK \cite{Usenko2017} queries a local circular buffer storing the field for real-time collision avoidance. Fast-Planner \cite{Zhou2019} and RAPTOR \cite{raptor} couple a kinodynamic front-end with a B-spline back-end driven by ESDF gradients.

Finally, the field has moved toward learned distance representations queried in a differentiable manner. iSDF \cite{Ortiz:etal:iSDF2022} trains a neural SDF online from posed depth images, and subsequent work has scaled the idea through hybrid grid-network representations and lightweight architectures for large scenes \cite{vasilopoulos_hiosdf_2024, shi2026}. In \cite{jacquet2025} the authors embed a local neural ESDF into a nonlinear MPC, using its analytical gradients as collision-avoidance constraints, and \cite{bukhari2025} proposes a composite neural ESDF combining static reconstructions with object-level representations for dynamic indoor scenes. Moreover, \cite{gil2026} introduces a framework that optimizes natural polynomial trajectories directly over a neural ESDF representation. These works establish that neural fields can be differentiated efficiently enough for online use, but they restrict the field to the immediate vicinity of the robot within a control horizon \cite{jacquet2025}, assume a reconstruction obtained beforehand \cite{bukhari2025}, or do not achieve a suitable planning rate and dynamic compliance for practical use \cite{gil2026}. We propose a mapping subsystem that avoids those limitations by being globally usable without prior reconstruction, as shown in Section \ref{sec:exper}, as well as a polynomial representation that achieves kinodynamic compliance with a suitable planning rate. 

% A second axis along which trajectory planners can be classified is how they encode environmental information. ESDF-based planners have flourished in the last decade leveraging the advantages explained in Section \ref{sec:intro}. Early continuous-time formulations such as \cite{Oleynikova2016} optimize a polynomial trajectory against an ESDF cost to enable online UAV replanning, while EWOK \cite{Usenko2017} represents the trajectory with uniform B-splines while querying a local 3D circular buffer storing the ESDF for real-time collision avoidance. Fast-Planner \cite{Zhou2019} couples a kinodynamic front-end with a B-spline back-end that takes advantage of ESDF gradients, and RAPTOR \cite{raptor} extends this scheme with perception-awareness, jointly optimizing trajectory and yaw against the ESDF to improve robustness in fast flight.

% More recently, the field has moved toward learned distance representations that can be queried in a differentiable manner. In \cite{jacquet2025} the authors embed a local neural ESDF into a nonlinear MPC formulation, using its analytical gradients as collision-avoidance constraints during online control. In \cite{bukhari2025} they go a step further and propose a differentiable composite neural ESDF that combines static scene reconstructions with object-level representations, which allows trajectory optimization in dynamic indoor environments where the underlying distance field changes over time.

\section{Framework structure} \label{sec:frame}

\begin{figure*}[t!]
\centerline{\includegraphics[width=.95\textwidth]{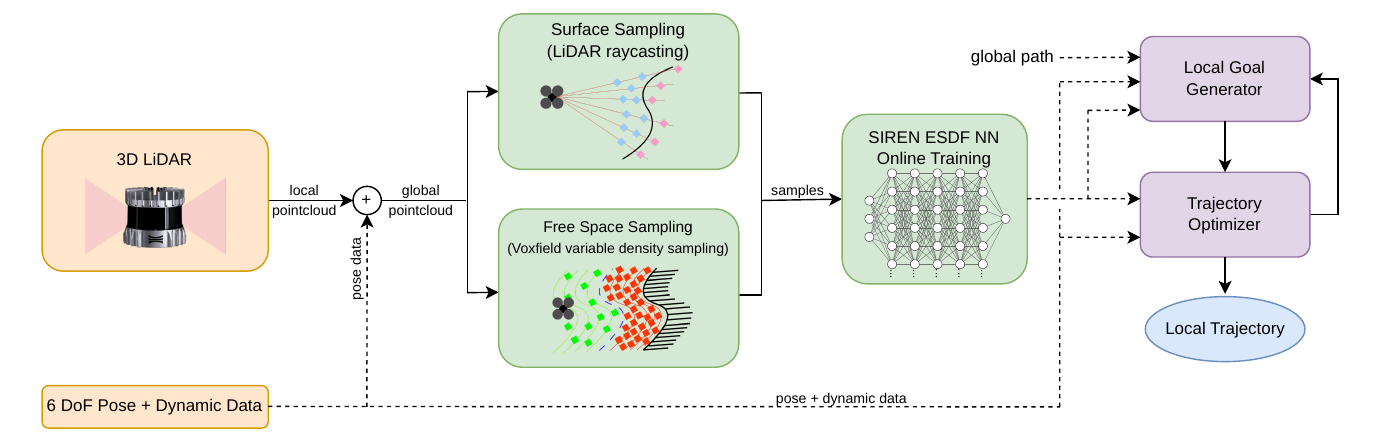}}
\vspace{-5mm}
\caption{Proposed framework overview. Sensor input (orange), neural ESDF generation (green) and local trajectory planning via non-linear optimization (purple). Surface samples from LiDAR raycasting and free-space samples with Voxfield's variable density train a SIREN network online, which is then queried as a continuous distance function in the optimizer. Solid arrows are the data pipeline; dashed arrows are queries and side inputs.}
\vspace{-5mm}
\label{architecture_fig}
\end{figure*}

Figure \ref{architecture_fig} shows a complete overview of the proposed trajectory planning pipeline. We can subdivide the framework into a neural mapping subsystem and a local trajectory planning subsystem.

\subsection{Online Mapping with a Global ESDF Neural Network}

Our framework's environment is represented by a neural ESDF trained online as the drone flies. An onboard 3D LiDAR provides point clouds that are transformed into global coordinates via global localization and converted into training samples by two complementary methods (Fig. \ref{architecture_fig}). The parameters shown in this section are defined and tuned in Section \ref{sec:exper}.

Surface sampling traces virtual rays from the LiDAR towards points of the latest cloud and estimates their ESDF against a FIFO buffer of the last $n_{\text{PCD}}$ point clouds, which attenuates LiDAR noise and prevents ESDF overestimation \cite{Ortiz:etal:iSDF2022, vasilopoulos_hiosdf_2024}. Pairs with $\left|ESDF_\text{k}\right| > t_{\text{t}}$ are discarded to avoid errors \cite{Ortiz:etal:iSDF2022}, and up to $n_{\text{sur}}$ of the remaining pairs are kept per training iteration.

Free space sampling covers regions near and far from obstacles using a modified Voxfield \cite{pan2022} node restricted to a low-resolution local map. Cropping keeps the message deserialization time bounded as the explored area grows and focuses the samples on the region relevant to local planning, as well as limiting the stored data memory usage. Samples are split into an obstacle zone ($ESDF_\text{k} < d_{\text{lim}}$) and a free zone ($ESDF_\text{k} > d_{\text{lim}}$), from which $\rho_{\text{obs}} \times j_{\text{obs}}$ and $\rho_{\text{free}} \times j_{\text{free}}$ samples are drawn. Choosing $\rho_{\text{obs}} > \rho_{\text{free}}$ concentrates network accuracy where trajectory safety is most critical.

Both sample sets feed a SIREN network \cite{sitzmann2020implicitneuralrepresentationsperiodic}, an architecture well suited to represent continuous scalar fields such as ESDFs \cite{vasilopoulos_hiosdf_2024, gil2026}. Training runs asynchronously for $n_{\text{epochs}}$ epochs on the most recent samples, so the planner always queries the latest available network at its own rate. 

\subsection{Local Trajectory Planning}

The planning loop, as a pure local planner, expects to have a precomputed global path in the form of successive 3D coordinates. To calculate an appropriate local goal, we define the local environment (local window from now on) as a user-configurable three-dimensional $L_x \times L_y \times L_z$ prismatic space centered on the drone.

The framework chooses the last coordinate of the global path within the local window as a provisional local goal $G'$. Assuming that a safe local goal $G$ should satisfy $ESDF(G) \geq ESDF_s$, where $ESDF_s$ is the minimal distance considered safe from the closest obstacle, this provisional goal may not meet this restriction. Let $S_{G'}(r)$ denote a certain spherical surface of radius $r$ centered at $G'$. The system performs parallel queries of the neural network of points $p \in S_{G'}(r)$ that land inside the local window for increasing values of $r$ (in $\Delta r$ steps), looking for the point closest to $G'$ such that $ESDF(p) \geq ESDF_s$. This point is then defined as the local goal $G=(x_G, y_G, z_G)$.

After that, the trajectory planning is performed using a non-linear optimizer that considers kinodynamic restrictions to calculate a suitable polynomial trajectory to the local goal. This continues until the drone reaches the global goal.

\section{Continuous Trajectory Planner on ESDFs} \label{sec:path}

This section describes the proposed trajectory planner based on a single non-linear optimizer using Ceres Solver\footnote{http://ceres-solver.org/}.

\subsection{Chebyshev representation and state vector}

%Before diving into the optimization process, it is important to first explain both the chosen trajectory representation and the state vector to be optimized. 
The purpose of this trajectory planner is to provide a continuous 3D trajectory for an omnidirectional drone as a parametric representation in the form $\mathbf{p}(t)=(x(t),y(t),z(t))$. Each component will be modeled as a fifth-order polynomial, combining the flexibility of high-order polynomials while maintaining a relatively low number of parameters to be optimized. 

The most common design choice for polynomial representation of 3D trajectories is B-splines. They provide formal guarantees of constraint satisfaction through the convex hull property, but this comes at the cost of a larger number of decision variables. More critically, even when B-splines are employed, jerk constraints remain fundamentally non-convex and computationally expensive to handle, often requiring specialized reformulation techniques to achieve convexity \cite{Ji2024Convex}. 

In this approach, we have chosen a representation based on Chebyshev polynomials; they offer a distinct advantage in the context of dynamic constraints and efficiency. As they are orthogonal, this choice is not merely a representational convenience. Recent work has demonstrated that, when using gradient-based optimizers, the Chebyshev basis consistently outperforms the classical polynomial formulation in terms of convergence behavior \cite{Waclawek2024MachineLO}. This superior convergence allows our optimizer to find high-quality solutions in fewer iterations. As we show in Section \ref{sec:exper}, this formulation improves optimizer performance compared to classical monomial power basis approaches \cite{gil2026}.
Additionally, a polynomial representation allows direct calculation of key characteristics of the final trajectory, such as its total length, velocity, or acceleration. This will be leveraged when defining the cost functions described throughout this section. 

The Chebyshev polynomials of the first kind, $T_n$, are defined by the following trigonometric expression.
\begin{equation}
% T_n (\cos\theta) = \cos(n\theta) 
T_n (s) = \cos\bigl(n \arccos(s)\bigr), \quad s \in [-1, 1] 
\end{equation}

\noindent There is also a recursive way to obtain them:
\begin{align} 
T_0 (s) &= 1 \\
T_1(s) &= s \\
T_{n+1}(s) &= 2sT_n(s) - T_{n-1}(s) \label{recursive}
\end{align}

The three fifth-order polynomials that represent the trajectory to be optimized can be written as follows:
\begin{align}
\begin{split}
x(s) ={}& c_{0}T_{5}(s) + c_{1}T_{4}(s) + c_{2}T_{3}(s) \\
        &+ c_{3}T_{2}(s) + c_{4}T_{1}(s) + c_{5}T_{0}(s)
\end{split}
\label{traj_eq_1} \\[0.5em]
\begin{split}
y(s) ={}& c_{6}T_{5}(s) + c_{7}T_{4}(s) + c_{8}T_{3}(s) \\
        &+ c_{9}T_{2}(s) + c_{10}T_{1}(s) + c_{11}T_{0}(s)
\end{split}
\label{traj_eq_2} \\[0.5em]
\begin{split}
z(s) ={}& c_{12}T_{5}(s) + c_{13}T_{4}(s) + c_{14}T_{3}(s) \\
        &+ c_{15}T_{2}(s) + c_{16}T_{1}(s) + c_{17}T_{0}(s)
\end{split}
\label{traj_eq_3}
\end{align}

\noindent where $c_k$ denotes coefficient $k$. Using the recursive definition (\ref{recursive}), we can obtain the remaining Chebyshev polynomials:
\begin{align} 
T_2 (s) &= 2s^2-1 \\
T_3(s) &= 4s^3-3s \\
T_4(s) &= 8s^4-8s^2+1 \\
T_5(s) &= 16s^5-20s^3+5s
\end{align}

We will be using the original $s\in[-1, 1]$ Chebyshev domain during the optimization, so that $s=-1$ corresponds to the local start and $s=1$ to the local goal. Using this normalized variable avoids ill-conditioned optimization problems and facilitates smooth convergence to the solution. The equivalence between the normalized variable $s$ and real local trajectory time $t\in[0,T]$, with $T$ the total local trajectory time, can be written as $s=2t/T - 1$.

We can then define the state vector $\mathbf{c}$ considered by the optimizer as the one that contains all eighteen coefficients of the polynomials and the total local trajectory time:
\begin{equation}
\mathbf{c} = 
\begin{bmatrix}
c_0, c_1, c_2, \dots, c_{15}, c_{16}, c_{17}, T
\end{bmatrix}
\label{eq:state}
\end{equation}

The trajectory represented by equations (\ref{traj_eq_1}-\ref{traj_eq_3}) refers to a local reference system, where the drone remains centered in the middle of the local window.

The initial state vector that is fed to the optimizer is the state vector of the last trajectory optimizer occurrence. As a local trajectory planning framework, both the local start and the local goal suffer minimal changes between adjacent planning occurrences. This allows every local trajectory output from the optimizer to be a good initial estimate of the next one, which leads to the system performing fewer optimization iterations, drastically reducing the total optimization time.

For the case where there is no previous occurrence of the planning loop, the framework uses an initial state vector representing a straight line at constant velocity (set to half the drone's maximum velocity $v_{\max}$) between the local start and local goal. This constitutes the optimal spatial solution in the absence of obstacles while also being fast to compute.

% There are only seven non-zero states in the resulting initial state, which can be calculated as follows.
% \begin{align*}
% c_4^i &= \frac{x_{\text{goal}} - x_{\text{start}}}{2}, &
% c_5^i &= \frac{x_{\text{goal}} + x_{\text{start}}}{2}, &
% c_{10}^i &= \frac{y_{\text{goal}} - y_{\text{start}}}{2} \\
% c_{11}^i &= \frac{y_{\text{goal}} + y_{\text{start}}}{2}, &
% c_{16}^i &= \frac{z_{\text{goal}} - z_{\text{start}}}{2}, &
% c_{17}^i &= \frac{z_{\text{goal}} + z_{\text{start}}}{2}
% \end{align*}
% \[
% T^i = \frac{2(\|\mathbf{p}_{\text{goal}}\|-\|\mathbf{p}_{\text{start}}\|)}{v_{\max}}
% \]

% The resulting first initial state $\mathbf{c^i}$ is then defined as the following vector.
% \begin{equation*}
% \mathbf{c^i} = 
% \begin{bmatrix}
% 0, 0, 0, 0, c_4^i, c_5^i, 0, 0, 0, 0, c_{10}^i, c_{11}^i, 0, 0, 0, 0, , c_{16}^i, c_{17}^i, T^i
% \end{bmatrix}
% \end{equation*} 

\subsection{Cost Functions}

A key feature of our framework is the ease with which constraints can be integrated through modular cost functions. The presented formulation comprises seven cost functions, each contributing one or more residuals that the optimizer seeks to minimize through an iterative process. The resulting objective function minimized by Ceres Solver is defined as
% \begin{equation}
% \min_{\mathbf{c}} \; \frac{1}{2} \sum_{i}\sum_{j} \left( \| e_{i,j}(\mathbf{c}) \|^2 \right)
% \end{equation}

% \noindent with $\mathbf{c}$ the state vector and $e_{i,j}$ the residual $j$ for the cost function $i$. In this subsection, we detail the constraints and their corresponding residuals. Every residual $e_{i,j}$ includes a weighting parameter $w_i$ that will be discussed in Section \ref{sec:exper} as part of the optimizer tuning process.

\begin{equation}
\min_{\mathbf{c}} \; \frac{1}{2} \sum_{i}\sum_{j} \left( \| w_i  e_{i,j}(\mathbf{c}) \|^2 \right)
\end{equation}

\noindent with $\mathbf{c}$ the state vector, $e_{i,j}$ the residual $j$ for the cost function $i$ and $w_i$ the weighting parameter for the cost function $i$. In this subsection, we detail the constraints and their residuals. The weighting factors will be discussed in Section \ref{sec:exper}.

Although the Chebyshev formulation improves optimizer performance, it can be difficult to follow. To address this, we will rearrange the equations (\ref{traj_eq_1}-\ref{traj_eq_3}) for better clarity, resulting in the following monomial basis formulation:
% \[
% \begin{aligned}
% p_{0x} &= c_5-c_3+c_1, &
% p_{1x} &= c_4-3c_2+5c_0, \\
% p_{2x} &= 2c_3-8c_1, &
% p_{3x} &= 4c_2-20c_0, \\
% p_{4x} &= 8c_1, &
% p_{5x} &= 16c_0
% \end{aligned}
% \]
% \[
% \begin{aligned}
% p_{0y} &= c_{11}-c_9+c_7, &
% p_{1y} &= c_{10}-3c_8+5c_6, \\
% p_{2y} &= 2c_9-8c_7, &
% p_{3y} &= 4c_8-20c_6, \\
% p_{4y} &= 8c_7, &
% p_{5y} &= 16c_6
% \end{aligned}
% \]
% \[
% \begin{aligned}
% p_{0z} &= c_{17}-c_{15}+c_{13}, &
% p_{1z} &= c_{16}-3c_{14}+5c_{12}, \\
% p_{2z} &= 2c_{15}-8c_{13}, &
% p_{3z} &= 4c_{14}-20c_{12}, \\
% p_{4z} &= 8c_{13}, &
% p_{5z} &= 16c_{12}
% \end{aligned}
% \]
\begin{align}
x(s) ={}& p_{5x}s^5 + p_{4x}s^4 + p_{3x}s^3 + p_{2x}s^2 + p_{1x}s + p_{0x}\label{traj_eq_1p} \\[0.5em]
y(s) ={}& p_{5y}s^5 + p_{4y}s^4 + p_{3y}s^3 + p_{2y}s^2 + p_{1y}s + p_{0y}\label{traj_eq_2p} \\[0.5em]
z(s) ={}& p_{5z}s^5 + p_{4z}s^4 + p_{3z}s^3 + p_{2z}s^2 + p_{1z}s + p_{0z}\label{traj_eq_3p}
\end{align}

\subsubsection{Total Length Cost Function}

A constraint that penalizes the total length of the local trajectory, leading to more efficient navigation.

The length $L$ of a certain parametric 3D curve $\mathbf{r(s)}=(x(s),y(s),z(s))$ in the range $[-1,1]$ can be calculated as follows.
\begin{equation}
    L = \int_{-1}^1 \lVert \mathbf{r}'(s) \rVert \, ds = \int_{-1}^1 \sqrt{
\left(\frac{dx}{ds}\right)^2 +
\left(\frac{dy}{ds}\right)^2 +
\left(\frac{dz}{ds}\right)^2
}\, ds
\label{long_arco}
\end{equation}

We use the Gauss--Legendre quadrature, which provides a way to approximate integrals and is stated as
\begin{equation}
    \int_{-1}^1 f(\xi)d\xi \approx \sum_{i=1}^n b_if(\xi_i)
    \label{legendre_eq}
\end{equation} 

\noindent with $\xi_i$ the roots of the degree-$n$ Legendre polynomial $P_n(\xi)$ and $b_i$ the associated weights.
 
%This approximation yields the exact value of the integral whenever $f(\xi)$ is a polynomial of degree $2n-1$ or less. %In our case, the derivatives $x'(\xi)$, $y'(\xi)$, and $z'(\xi)$ are polynomials of degree 4, so the radicand in $f(\xi)=\sqrt{x'(\xi)^2+y'(\xi)^2+z'(\xi)^2}$ is a polynomial of degree 8. 
Although the square root makes the integrand non-polynomial, for which the method theoretically yields the exact value, Gauss--Legendre quadrature still converges rapidly for smooth integrands. We choose $n=5$, which we found sufficiently accurate in practice.

%Since the integration limits in \eqref{long_arco} and \eqref{legendre_eq} coincide, no change of variables is required. 
With this method, the residual of this cost function reduces to a weighted sum of the norm of the trajectory derivatives evaluated at the quadrature nodes:
\begin{equation}
    e_{1,1}(\mathbf{c})=L = \sum_{i=1}^5 b_i\sqrt{x'(\xi_i)^2+y'(\xi_i)^2+z'(\xi_i)^2}
\end{equation}

The derivatives for equations (\ref{traj_eq_1}-\ref{traj_eq_3}) are trivial. Computing the nodes ($\xi_i$) and weights ($b_i$) of the Gauss--Legendre quadrature rule is beyond the scope of this paper; we refer the reader to the classical algorithm by Golub and Welsch ~\cite{golub1969calculation}. Table \ref{gauss_legendre_table} lists the values for the 5-point rule. \\

\begin{table}[!tb]
\centering
\caption{Nodes and weights for the 5-point Gauss-Legendre quadrature approximation}
\begin{tabular}{c c c}
\hline
$i$ & $\xi_i$ & $b_i$ \\
\hline
$1$ & $0$ & $\frac{128}{225}$ \\
$2$, $3$ & $\pm\frac{1}{3}\sqrt{5 - 2 \sqrt{10/7}}$ & $\frac{322 + 13 \sqrt{70}}{900}$ \\
$4$, $5$ & $\pm\frac{1}{3}\sqrt{5 + 2 \sqrt{10/7}}$ & $\frac{322 - 13 \sqrt{70}}{900}$ \\\\
\hline
\end{tabular}
\label{gauss_legendre_table}
\end{table}

\subsubsection{ESDF Cost Function}

A constraint that integrates the ESDF information into the optimizer, avoiding trajectories close to obstacles. The framework uses the Ceres Solver evaluation callback to query the ESDF value and the gradient of a set of $n_{\text{ESDF}}$ points on the trajectory to the closest obstacle in every optimizer iteration. The samples are taken in an equidistributed manner along the $s$ axis, avoiding the fixed local start and goal. The residual of this function is calculated as follows.
\begin{equation*} \label{ESDF_cost}
e_{2,1}(\mathbf{c}) = \frac{1}{n_{\text{ESDF}}} \sum_{k=0}^{n_{\text{ESDF}}-1} e^{-\alpha(o_k - \sigma)}
\end{equation*}

It is designed to be predominant when the ESDF distance $o_k$ of any sample $k$ falls below a certain threshold $\sigma$. The parameter $\alpha$ defines how aggressive the function behaves. Both $\alpha$ and $\sigma$ are further explored in Section \ref{sec:exper}.\\

\subsubsection{Smoothness Cost Function}

A constraint that incentivizes the optimizer to increase low-order coefficients, leading to smooth velocity and acceleration profiles. We define twelve residuals for this cost function.
\[
\begin{aligned}
e_{3,1}(\mathbf{c}) &= p_{2x}, &
e_{3,2}(\mathbf{c}) &= p_{3x}, \\
e_{3,3}(\mathbf{c}) &= h_{\text{stiff}} \, p_{4x}, &
e_{3,4}(\mathbf{c}) &= h_{\text{stiff}} \, p_{5x}
\end{aligned}
\]
\[
\begin{aligned}
e_{3,5}(\mathbf{c}) &= p_{2y}, &
e_{3,6}(\mathbf{c}) &= p_{3y}, \\
e_{3,7}(\mathbf{c}) &= h_{\text{stiff}} \, p_{4y}, &
e_{3,8}(\mathbf{c}) &= h_{\text{stiff}} \, p_{5y}
\end{aligned}
\]
\[
\begin{aligned}
e_{3,9}(\mathbf{c}) &= z_{\text{stiff}} \, p_{2z}, &
e_{3,10}(\mathbf{c}) &= z_{\text{stiff}} \, p_{3z}, \\
e_{3,11}(\mathbf{c}) &= z_{\text{stiff}} \, h_{\text{stiff}} \, p_{4z}, &
e_{3,12}(\mathbf{c}) &= z_{\text{stiff}} \, h_{\text{stiff}} \, p_{5z}
\end{aligned}
\]

The weighting factor $h_{\text{stiff}}$ further increases the weighting of the fourth- and fifth order coefficients. We use the additional weighting factor $z_{\text{stiff}}$ as a result of the practical limitations of a 3D LiDAR. This sensor captures a point cloud with a higher density at the current flight altitude. This results in slightly better network performance around this altitude. Therefore, the weighting encourages the optimizer to prioritize longitudinal movements over vertical movements, which leverages the properties of the zones of the neural network that are better tuned. \\

\subsubsection{Fixed Start and End Cost Function}

A constraint that ensures that the previously computed local start and local goal are preserved through the optimization process.
\[
\begin{aligned}
\mathbf{e}_{4,1}(\mathbf{c}) &= \Bigl(\mathbf{p}(-1)-\mathbf{p}_{\text{start}}\Bigr) \\
\mathbf{e}_{4,2}(\mathbf{c}) &= \Bigl(\mathbf{p}(1)-\mathbf{p}_{\text{goal}}\Bigr)
\end{aligned}
\]

As Ceres Solver does not support complex multi-state hard constraints, they are introduced as cost functions with a high weighting factor. \\

\subsubsection{Initial Dynamic State Cost Function}

A constraint that ensures that the drone's velocity and acceleration at the local start match its current values. Let $\mathbf{v}_{\text{c}}$ be the current velocity of the drone and $\mathbf{a}_{\text{c}}$ be its current acceleration. We can then define the residuals as follows:
\[
\begin{aligned}
\mathbf{e}_{5,1}(\mathbf{c}) &= \Bigl(\mathbf{p}'(-1)-\mathbf{v}_{\text{c}}\frac{T}{2}\Bigr) \\
\mathbf{e}_{5,2}(\mathbf{c}) &= \Bigl(\mathbf{p}''(-1)-\mathbf{a}_{\text{c}}\frac{T^2}{4}\Bigr)
\end{aligned}
\]

% \subsubsection{Goal Dynamic State Cost Function}

% A soft constraint that allows the user to encourage a certain desired dynamic state at the local goal. Let $\mathbf{v}_{\text{d}}$ and $\mathbf{a}_{\text{d}}$ be the desired velocity and acceleration at the local goal, the residuals can be written as follows:
% \[
% \begin{aligned}
% \mathbf{e}_{6,1}(\mathbf{c}) &= \Bigl(\mathbf{p}'(1)-\mathbf{v}_{\text{d}}\frac{T}{2}\Bigr) \\
% \mathbf{e}_{6,2}(\mathbf{c}) &= \Bigl(\mathbf{p}''(1)-\mathbf{a}_{\text{d}}\frac{T^2}{4}\Bigr)
% \end{aligned}
% \]

\subsubsection{Dynamic Limits Enforcement Cost Function}

A constraint that enforces the drone's velocity, acceleration, and jerk limits by penalizing dynamic states that exceed these limits at $n_{\text{DYN}}$ uniformly distributed sample points along the trajectory, using softplus functions that approximate the behavior of a hard constraint while remaining differentiable. Let $v_{\max}$, $a_{\max}$, and $j_{\max}$ denote the maximum allowed velocity, acceleration, and jerk of the drone, respectively. The residuals are defined for each sample point $i \in [0, n_{\text{DYN}}-1]$ as follows:
\[
\begin{aligned}
e_{6,3i+1}(\mathbf{c}) &= \frac{1}{n_{\text{DYN}}}\,\text{softplus}_{k_{\text{softplus}}}\left(\frac{\frac{2}{T}\|\mathbf{p}'(s_i)\|}{v_{\max}} - 1\right) \\
e_{6,3i+2}(\mathbf{c}) &= \frac{1}{n_{\text{DYN}}}\,\text{softplus}_{k_{\text{softplus}}}\left(\frac{\frac{4}{T^2}\|\mathbf{p}''(s_i)\|}{a_{\max}} - 1\right) \\
e_{6,3i+3}(\mathbf{c}) &= \frac{1}{n_{\text{DYN}}}\,\text{softplus}_{k_{\text{softplus}}}\left(\frac{\frac{8}{T^3}\|\mathbf{p}'''(s_i)\|}{j_{\max}} - 1\right)
\end{aligned}
\]
\noindent where $k_{\text{softplus}}$ is the softplus sharpness constant. \\

\subsubsection{Trajectory Time Minimization Cost Function}

A constraint that penalizes trajectory time, leading to efficient navigation. The residual can be written as follows:

\begin{equation*} \label{time_cost}
e_{7,1}(\mathbf{c}) = T
\end{equation*}

A lower bound is also imposed on $T$ at each trajectory planning occurrence, corresponding to the time required to traverse the straight-line distance between the local start and goal at $v_{\max}$, preventing the optimizer from exploring physically infeasible values of $T$.

\section{Experimental Validation and Benchmarking} \label{sec:exper}

To assess the performance of the proposed framework, in this section we benchmark it against two state-of-the-art local trajectory planners: the Fast-Planner \cite{Zhou2019} and the EGO-Planner \cite{ego-planner}. The former relies on a discrete ESDF built incrementally from an occupancy map. We use the topological branch as it is their best recognized version of the framework. The latter eliminates the need for maintaining an ESDF altogether, as collision information is estimated only for trajectory segments that intersect obstacles, and the gradient used for obstacle avoidance is obtained directly from the occupancy grid map.

\begin{figure}[t]
    \centering
    \includegraphics[width=0.8\columnwidth]{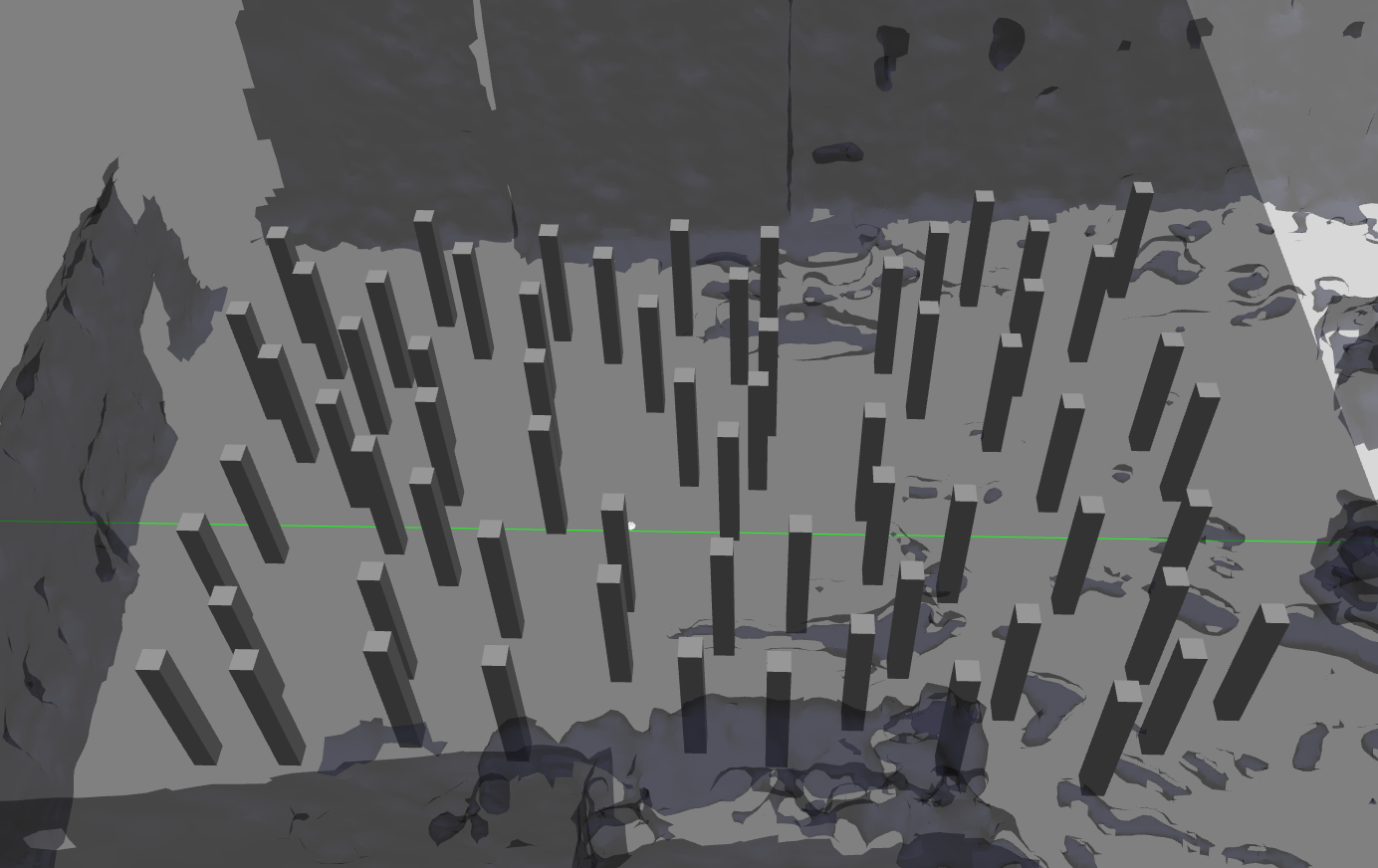}
    \caption{Simulated cluttered scenario. Columns are unmapped and thus unknown to the global path planner.}
    \label{fig:maps}
\end{figure}

The benchmarking is conducted in a $40\,\mathrm{m} \times 65\,\mathrm{m} \times 7.5\,\mathrm{m}$ simulated environment populated with square pillars of cross-section $0.5\,\mathrm{m} \times 0.5\,\mathrm{m}$, randomly distributed with a density of $0.15\,\mathrm{pillars/m^2}$, resulting in a highly cluttered scenario, similar to the one used in \cite{ego-planner}. Figure \ref{fig:maps} shows the simulated map. All frameworks are evaluated on the same hardware: an AMD Ryzen 7 7735HS CPU (3.2\,GHz base clock) with 32\,GB of RAM, running Ubuntu 20.04 LTS inside a Docker container over Ubuntu 22.04, with a RTX 4060 Laptop 8GB RAM GPU.

% \begin{figure}[t]
%     \centering
%     \includegraphics[width=\columnwidth]{Figures/Validation/gazebo_map.png}
%     \caption{Simulated cluttered scenario}
%     \label{fig:gazebo_map}
% \end{figure}

% \begin{figure*}[t]
%     \centering
%     \includegraphics[width=0.475\textwidth]{Figures/Validation/gazebo_map.png}
%     \hfill
%     \includegraphics[width=0.51\textwidth]{Figures/Validation/rviz_map.png}
%     \caption{Simulated cluttered scenario (left) and example LiDAR pointcloud (right). Columns are unmapped and thus unknown to the global path planner.}
%     \label{fig:maps}
% \end{figure*}

\subsection{Parameter and weight tuning}

Our framework relies on a set of parameters whose optimal values are inherently application-dependent. This section shows the values used throughout the experimental validation, summarized in Table~\ref{tab:params}, together with the rationale behind their selection.

The mapping constants were chosen to trade off accuracy against training time. The cost-function weights were tuned incrementally, introducing one constraint at a time and adjusting its weight until the desired behavior was obtained.

The remaining parameters were obtained from previous studies ($\alpha$ and $\sigma$, following the analysis in \cite{gil2025, gil2026}), derived from the dimensions of the local window ($n_{\mathrm{ESDF}}$ and $n_{\mathrm{DYN}}$), tuned experimentally ($h_{\mathrm{stiff}}$, $z_{\mathrm{stiff}}$ and $k_{\mathrm{softplus}}$), or set to typical values for commercial platforms ($v_{\max}$, $a_{\max}$ and $j_{\max}$).

\begin{table}[!tb]
\centering
\footnotesize
\setlength{\tabcolsep}{4pt}
\caption{Parameters used in the experimental validation.}
\label{tab:params}
\begin{tabular}{@{}llrl@{}}
\hline
Symbol & Description & Value & Unit \\
\hline
\multicolumn{4}{@{}l}{\textit{Mapping}} \\
$n_{\mathrm{PCD}}$     & PCDs for ESDF comparison     & $3$              & -- \\
$t_{\mathrm{t}}$       & Sample truncation distance   & $0.2$            & m \\
$n_{\mathrm{sur}}$     & Surface sampling points      & $5\times10^{3}$  & -- \\
$d_{\mathrm{lim}}$     & ESDF limit between zones     & $1.5$            & m \\
$\rho_{\mathrm{obs}}$  & Point density, obstacle zone & $0.05$           & -- \\
$\rho_{\mathrm{free}}$ & Point density, free zone     & $0.01$           & -- \\
$n_{\text{epochs}}$ & Epochs per iteration     & $10$           & -- \\
\hline
\multicolumn{4}{@{}l}{\textit{Local window and goal selection}} \\
$L_{x}\!\times\!L_{y}\!\times\!L_{z}$ & Local window dimensions & $6.0\!\times\!6.0\!\times\!2.0$ & m \\
$\mathrm{ESDF}_s$ & Min. safe clearance for $G$ & $1.5$  & m \\
$\Delta r$        & Radial step for $G$ search  & $0.1$ & m \\
\hline
\multicolumn{4}{@{}l}{\textit{Cost-function weights}} \\
$w_1$ & Total length                       & $2\times10^{2}$ & -- \\
$w_2$ & ESDF barrier                       & $3\times10^{3}$ & -- \\
$w_3$ & Smoothness                         & $2$             & -- \\
$w_4$ & Fixed start and end                & $10^{5}$        & -- \\
$w_5$ & Initial dynamic state              & $10^{5}$        & -- \\
% $w_6$ & Goal dynamic state               & $10^{2}$        & -- \\
$w_6$ & Dynamic limits                     & $2\times10^{6}$        & -- \\
$w_7$ & Trajectory time                    & $5\times10^{3}$ & -- \\
\hline
\multicolumn{4}{@{}l}{\textit{Optimizer}} \\
$\alpha$                & ESDF cost sharpness   & $4$   & m$^{-1}$ \\
$\sigma$                & ESDF cost threshold   & $1.5$ & m \\
$k_{\mathrm{softplus}}$ & Softplus sharpness    & $50$  & -- \\
$n_{\mathrm{ESDF}}$     & ESDF sample points    & $20$  & -- \\
$n_{\mathrm{DYN}}$      & Dynamic sample points & $50$  & -- \\
$h_{\mathrm{stiff}}$    & High-order stiffness  & $50$  & -- \\
$z_{\mathrm{stiff}}$    & Vertical stiffness    & $100$ & -- \\
\hline
\multicolumn{4}{@{}l}{\textit{Platform limits}} \\
$v_{\max}$ & Maximum velocity     & $3.0$ & m/s \\
$a_{\max}$ & Maximum acceleration & $2.0$ & m/s$^2$ \\
$j_{\max}$ & Maximum jerk         & $4.0$ & m/s$^3$ \\
\hline
%\multicolumn{4}{@{}p{0.95\columnwidth}}{\scriptsize $^{\dagger}$ Theoretical hard constraint, implemented as a heavily weighted cost term.} \\
\end{tabular}
\end{table}

\subsection{Mapping Performance}

On the mapping side, we evaluate the cost of building and maintaining each environment representation---the neural implicit map in our case, the incrementally updated ESDF in Fast-Planner, and the occupancy grid map in EGO-Planner---in terms of memory footprint and update time, with a simulated LiDAR pointcloud as the main source of spatial information. Table \ref{tab:comp_memory} shows the update time and memory consumption of the three considered frameworks with their recommended resolutions.

\begin{table}[!tb]
\centering
\footnotesize
\setlength{\tabcolsep}{6pt}
\caption{Computation time and memory consumption of the spatial representation, averaged over 50 iterations.}
\label{tab:comp_memory}
\begin{tabular}{@{}lrr@{}}
\hline
Framework (resolution) & Update Time & Memory Consumption \\
\hline
Fast-Planner \cite{Zhou2019} ($0.1$ m) & $26.6$\,ms  & $2082.8$\,MB \\
EGO-Planner \cite{ego-planner} ($0.1$ m) & $6.1$\,ms & $334.7$\,MB \\
Ours & $609.1$\,ms & $1.0$\,MB \\
\hline
\end{tabular}
\end{table}

It can be seen that explicit volumetric structures of Fast-Planner and EGO-Planner are updated locally over a fixed set of voxels and are therefore fast, but their memory cost grows cubically with the mapped volume and with the inverse of the voxel size. Our representation reverses this balance, reducing the memory footprint at the cost of a slower update caused by the network training pipeline. More importantly, this cost is governed by the network architecture rather than the extent of the mapped space, thus requiring only a rough approximation of the map dimensions before the flight to adjust the network size.

Regarding the accuracy of the learned representation, we assessed it by comparing the queried distances and gradients with the ground-truth ESDF over training iterations. The network recovers the distance field with an error below $0.1 m$ for $83.3\%$ of the evaluated points,  below $0.3 m$ for $91.2\%$ and below $0.5$ m for $97.3\%$. The iterative nature of the optimization process mitigates the impact of the few incorrectly predicted points, resulting in safe navigation as long as the safety distance is correctly tuned. 

Although the optimizer only evaluates $n_\text{ESDF}$ discrete points along the trajectory, the spacing between samples is designed to be much smaller than the ESDF threshold in the current configuration. This, combined with the accuracy of the representation, makes it impossible for obstacles between samples to be overlooked by the planner.

Regarding the gradient, over $92.5\%$ of the samples exhibited the correct direction together with a magnitude within $[0.85, 1.15]$, consistent with the eikonal condition \cite{gropp2020implicit}. This is relevant as the optimizer relies on this gradient to steer the trajectory away from obstacles. The accurate gradient with near-unit norm allows for well-conditioned updates.

\subsection{Trajectory Planning Performance}

\begin{figure}[t]
    \centering
    \includegraphics[width=0.8\columnwidth]{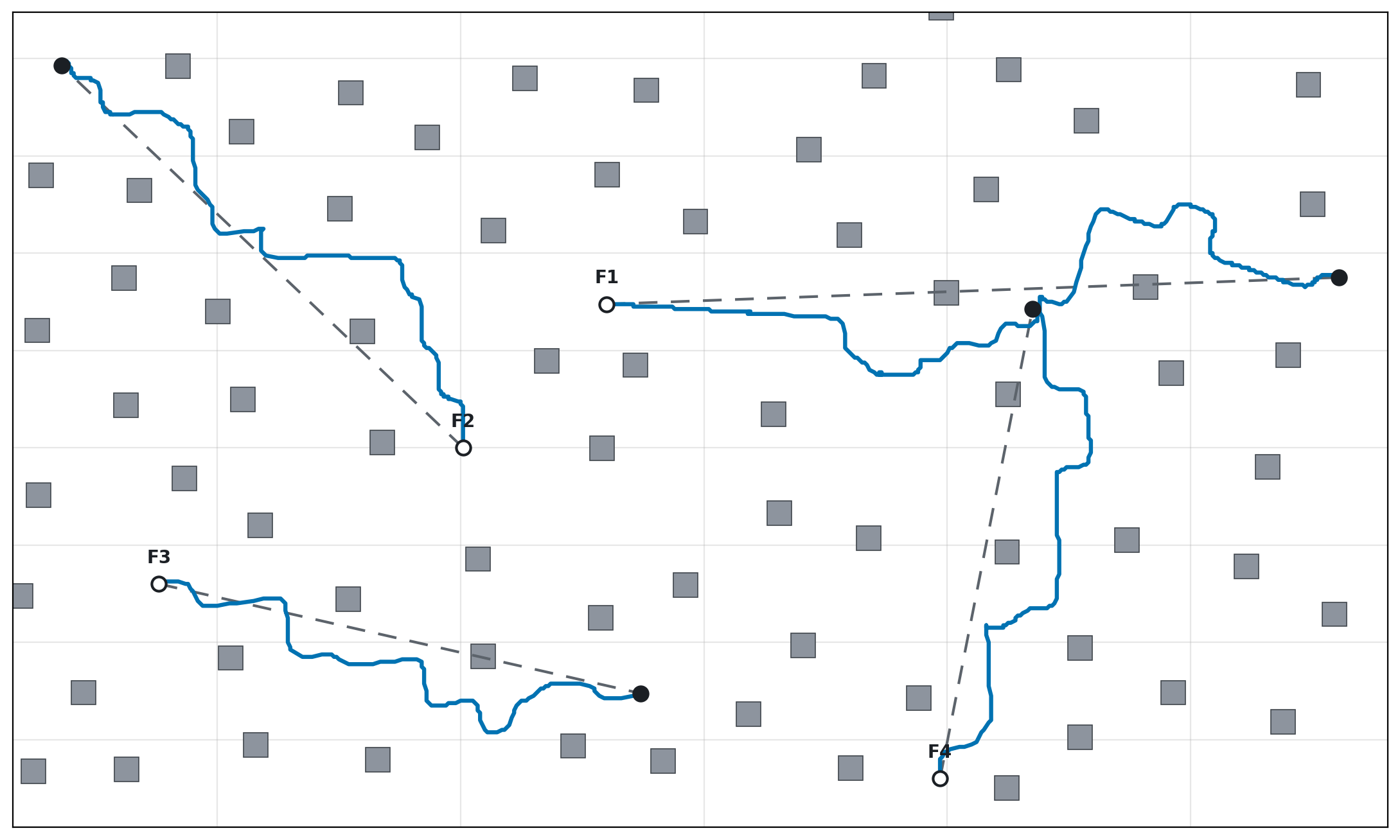}
    \caption{Top view of the simulated flights (F1 to F4), showing the pre-calculated column-unaware global paths (dashed black lines) and the manually controlled flight paths (continuous blue lines).}
    \label{fig:flown_paths}
\end{figure}

\begin{figure*}[t]
    \centering
    \includegraphics[width=.8\textwidth]{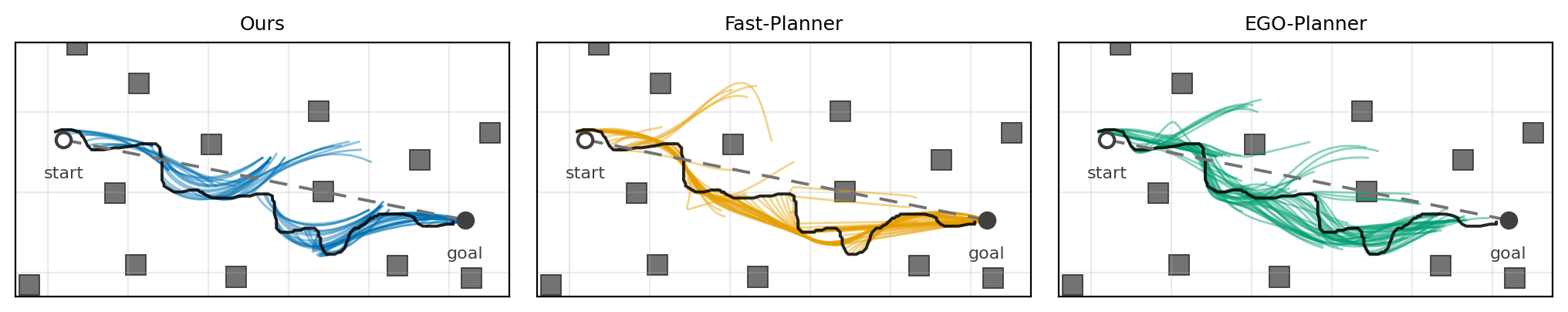}
    \caption{Top view of generated local trajectories for each benchmarked method on the F3 simulated flight. Only one out of every ten generated trajectories is shown, for clarity.}
    \label{fig:f3_flights}
\end{figure*}

Evaluating our trajectory optimizer against the other two state-of-the-art methods requires a fair evaluation protocol. To account for structural differences between frameworks, four flights along manually controlled paths were recorded in simulation together with all sensor data, as shown in Figure \ref{fig:flown_paths}. Then, the recordings were replayed to each of the three frameworks in turn with the same replanning rate, achieving over 1500 local trajectories under the same conditions for each benchmarked method. Figure \ref{fig:f3_flights} shows a portion of the local trajectories for one of the four flown paths. %Figure \ref{fig:flown_paths} shows the four paths taken across the simulated map and the precalculated column-unaware global paths associated with them. %The manually flown paths aim to create different situations along the map for the planners to solve in equal condition. %Figure \ref{fig:traj_compare} shows examples of different local trajectory outputs by the three benchmarked planners under the same conditions.

This procedure allows the behavior of the benchmarked frameworks to be compared under identical inputs across a wide set of controlled situations in the previously described cluttered environment. Since neither Fast-Planner nor EGO-Planner is able to displace its local goal away from obstacles, and both are designed to operate with large local ranges, their local horizon was extended to $6\mathrm{m}$ and the resulting trajectory was cropped to the horizon defined in Table~\ref{tab:params}, so that the total length and duration of the trajectory remain comparable.

Table~\ref{tab:benchmark} shows the measured results for all the output trajectories of the three planners. Both our framework and Fast-Planner returned a trajectory in every planning iteration, whereas EGO-Planner failed to reach a solution in $17.3\%$ of them. Our planner keeps the required $0.4$~m clearance in $99.2\%$ of the iterations and attains the highest mean and median distance to nearby obstacles, $1.119$~m and $1.028$~m respectively.

When comparing dynamic feasibility, it is only fair to check velocity and acceleration limits, as jerk is not enforced in the chosen state-of-the-art methods. Our framework complies with the desired dynamic limits in $99.4\%$ of the planning iterations. It achieves this behavior due to the dynamic limits being embedded in the optimizer costs as heavily weighted restrictions. The other methods achieve only $22.5\%$ and $52.1\%$ of dynamic compliance under the per-axis criterion they themselves use \cite{Zhou2019, ego-planner}. This follows from two design choices: feasibility enters their objective as a soft term weighted an order of magnitude below smoothness and path fitting, and the time reallocation triggered upon dynamic limits violation is applied once and re-optimized again without a second feasibility check. Furthermore, as stated above, none of them implement jerk limits. Our framework not only complies with the imposed limits, but shapes the trajectory so as to exploit them. This situates the median peak values of $|v|$, $|a|$ and $|j|$ within $4\%$ of their respective bounds.

Regarding computation time, our framework solves each planning iteration in $73.05$~ms on average, i.e. an update rate of $13.7$~Hz. Although this is one order of magnitude slower than the state-of-the-art methods, it remains well above the rate required for local planning. The table also shows that, while our framework generates safer trajectories than the other methods, the resulting path length remains comparable.

For the duration comparison to be fair, every trajectory exceeding the imposed limits was time-scaled by the smallest factor that brings velocity, acceleration and jerk back into range, so that only trajectories the drone could actually fly are compared. It should be taken into account that the jerk is included here, unlike in the feasibility analysis. Compliance measures whether a planner meets the constraints it models, whereas duration measures physical executability, for which all three limits must apply. The data show that our framework generates flight trajectories that take advantage of the drone's dynamic capabilities to calculate the fastest trajectories overall of the benchmarked methods.

\begin{table}[!tb]
\centering
\footnotesize
\setlength{\tabcolsep}{6pt}
\caption{Local planner comparison in cluttered simulated map, pooled over 4 independent flights. Dynamic limits set to $|v|\leq3$~$\mathrm{m/s}$, $|a|\leq2$~$\mathrm{m/s^2}$, $|j|\leq4$~$\mathrm{m/s^3}$.}
\label{tab:benchmark}
\begin{tabular}{@{}lrrr@{}}
\hline
Metric & Ours & Fast \cite{Zhou2019} & EGO \cite{ego-planner} \\
\hline
\multicolumn{4}{@{}l}{\textit{Safety}} \\
\quad Planning Success (\%) & $\mathbf{100.0}$ & $\mathbf{100.0}$ & $82.7$ \\
\quad Clearance $\geq0.4$~m (\%) & $99.2$ & $\mathbf{99.6}$ & $82.7$ \\
\quad Clearance, mean (m) & $\mathbf{1.119}$ & $0.967$ & $0.861$ \\
\quad Clearance, median (m) & $\mathbf{1.028}$ & $0.806$ & $0.667$ \\
\hline
\multicolumn{4}{@{}l}{\textit{Direct output dynamic feasibility ($v$ and $a$)}} \\
\quad Feasible, 3-D norm (\%) & $\mathbf{97.4}$ & $17.8$ & $33.7$ \\
\quad Feasible, per-axis (\%) & $\mathbf{99.4}$ & $22.5$ & $52.1$ \\
\quad $\max|v|$, median (m/s) & $\mathbf{2.896}$ & $2.812$ & $2.343$ \\
\quad $\max|v|$, worst (m/s) & $\mathbf{3.030}$ & $5.507$ & $3.963$ \\
\quad $\max|a|$, median (m/s$^2$) & $\mathbf{1.937}$ & $3.525$ & $2.332$ \\
\quad $\max|a|$, worst (m/s$^2$) & $\mathbf{2.138}$ & $13.949$ & $13.282$ \\
\quad $\max|j|$, median (m/s$^3$) & $\mathbf{3.905}$ & $12.705$ & $8.691$ \\
\quad $\max|j|$, worst (m/s$^3$) & $\mathbf{4.113}$ & $80.489$ & $115.652$ \\
\hline
\multicolumn{4}{@{}l}{\textit{Computation time}} \\
\quad Median (ms) & $37.82$ & $1.46$ & $\mathbf{0.98}$ \\
\quad Mean (ms) & $73.05$ & $7.24$ & $\mathbf{6.50}$ \\
\hline
\multicolumn{4}{@{}l}{\textit{Trajectory quality}} \\
\quad Path length (m) & $3.166$ & $\mathbf{3.060}$ & $3.197$ \\
\quad Executable duration (s) & $\mathbf{2.528}$ & $2.652$ & $2.779$ \\
\hline
\end{tabular}
\end{table}

\section{Conclusion} \label{sec:conc}

We present a non-linear optimization-based trajectory planner that uses a continuously updated neural ESDF with low memory usage. The results show improvements over state-of-the-art approaches in trajectory duration and safety, while maintaining a safe planning rate. It is also shown that the resulting trajectories exploit the drone's dynamic envelope rather than merely staying inside it.

The wide set of kinodynamic restrictions used during the optimization, as well as using trajectory time as an optimizable variable, has proven to yield a significant improvement in safety and dynamic compliance when compared to other methods. The online neural mapping used for environmental representation allows for reduced memory footprint, at the cost of longer map refresh times, without losing accuracy in a significant way.

As future work, reducing the computation time of the optimizer remains a promising investigation line. Another approach should be integrating other recent online mapping technologies to lower the map refresh latency while still remaining low-memory.

% \section*{Acknowledgment}

% This work was supported by the grants COBUILD (PID2024-161069OB-C31, funded by the Ministry of Science, Innovation and Universities, the Spanish Research Agency, and the European Regional Development Fund, MICIU /AEI /10.13039/501100011033 / FEDER, UE), and PICRAH 4.0 (PLEC2023-010353, funded by the Spanish Research Agency and the Ministry of Science, Innovation and Universities, MCIN /AEI /10.13039/501100011033)

% To balance between two columns the last page
\balance

\bibliographystyle{IEEEtran}
\bibliography{ral2026}

\end{document}